\documentclass[letterpaper]{article} %
\usepackage{aaai2026}  %
\usepackage{times}  %
\usepackage{helvet}  %
\usepackage{courier}  %
\usepackage[hyphens]{url}  %
\usepackage{graphicx} %
\usepackage{natbib}  %
\usepackage{caption} %
\usepackage{algorithm}
\usepackage{algorithmic}
\usepackage{amsmath}
\usepackage{amssymb}
\usepackage{booktabs}
\usepackage{dsfont}
\usepackage{multirow}
\usepackage{subcaption}
\usepackage{tabularx}
\usepackage{xspace}

\newcommand{\pointstyle}{\raisebox{-1.5pt}{\includegraphics[height=1.05em]{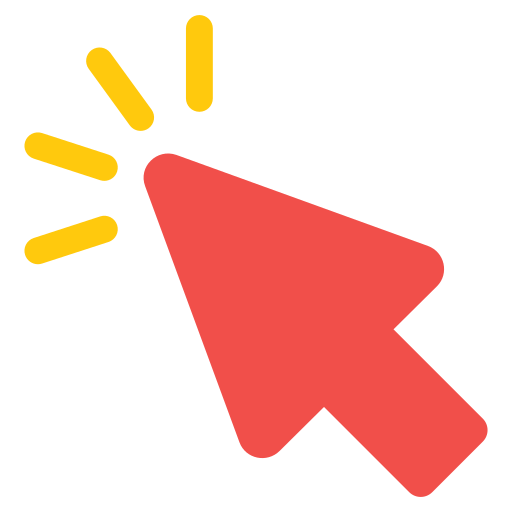}}\xspace}
\newcommand{\bboxstyle}{\raisebox{-1.5pt}{\includegraphics[height=1.05em]{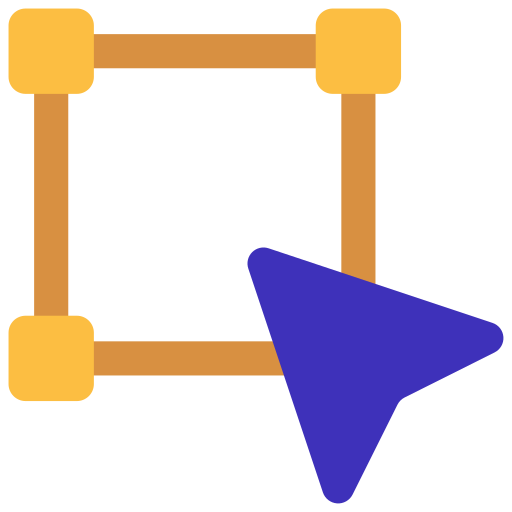}}\xspace}
\usepackage{newfloat}
\usepackage{listings}
\DeclareCaptionStyle{ruled}{labelfont=normalfont,labelsep=colon,strut=off} %
\floatstyle{ruled}
\newfloat{listing}{tb}{lst}{}
\floatname{listing}{Listing}
\title{Test-Time Reinforcement Learning for GUI Grounding via Region Consistency}
\author {
    Yong Du\textsuperscript{\rm 1,\rm 2,}\equalcontrib,
    Yuchen Yan\textsuperscript{\rm 1,}\equalcontrib,
    Fei Tang\textsuperscript{\rm 1},
    Zhengxi Lu\textsuperscript{\rm 1},\\
    Chang Zong\textsuperscript{\rm 3},
    Weiming Lu\textsuperscript{\rm 1,$\dagger$},
    Shengpei Jiang\textsuperscript{\rm 4},
    Yongliang Shen\textsuperscript{\rm 1,}\footnote{Corresponding author.}
}
\affiliations {
    \textsuperscript{\rm 1}Zhejiang University,\\
    \textsuperscript{\rm 2}Central South University,\\
    \textsuperscript{\rm 3}Zhejiang University of Science and Technology,\\
    \textsuperscript{\rm 4}SF Technology\\
    duyong@csu.edu.cn, \{yanyuchen, syl\}@zju.edu.cn
}

\begin{document}

\maketitle

\begin{abstract}
Graphical User Interface (GUI) grounding, the task of mapping natural language instructions to precise screen coordinates, is fundamental to autonomous GUI agents. While existing methods achieve strong performance through extensive supervised training or reinforcement learning with labeled rewards, they remain constrained by the cost and availability of pixel-level annotations. 
We observe that when models generate multiple predictions for the same GUI element, the spatial overlap patterns reveal implicit confidence signals that can guide more accurate localization.
Leveraging this insight, we propose \textbf{GUI-RC (\underline{R}egion \underline{C}onsistency)}, a test-time scaling method that constructs spatial voting grids from multiple sampled predictions to identify consensus regions where models show highest agreement. Without any training, GUI-RC improves accuracy by 2-3\% across various architectures on ScreenSpot benchmarks.
We further introduce \textbf{GUI-RCPO (\underline{R}egion \underline{C}onsistency \underline{P}olicy \underline{O}ptimization)}, transforming these consistency patterns into rewards for test-time reinforcement learning. By computing how well each prediction aligns with the collective consensus, GUI-RCPO enables models to iteratively refine their outputs on unlabeled data during inference. 
Extensive experiments demonstrate the generality of our approach: using only 1,272 unlabeled data, GUI-RCPO achieves 3-6\% accuracy improvements across various architectures on ScreenSpot benchmarks.
Our approach reveals the untapped potential of test-time scaling and test-time reinforcement learning for GUI grounding, offering a promising path toward more data-efficient GUI agents.
\end{abstract}

\begin{links}
    \link{Code}{https://github.com/zju-real/gui-rcpo}
    \link{Project}{https://zju-real.github.io/gui-rcpo}
\end{links}

\section{Introduction}

The rapid advancement of GUI (Graphical User Interface) agents is fundamentally transforming human-device interaction, enabling users to control complex interfaces in digital devices through natural language across diverse applications~\citep{gou2024navigatingdigitalworldhumans, tang2025surveymllmbasedguiagents, wang2024mobileagentautonomousmultimodalmobile}. At the heart of these systems lies GUI grounding, the critical capability to accurately map natural language instructions to precise pixel coordinates on UI elements~\citep{cheng2024seeclickharnessingguigrounding, tang2025thinktwiceclickonce}. This fundamental task determines whether an agent can successfully execute user commands, making it the cornerstone of reliable GUI automation.

Current approaches to GUI grounding have achieved impressive results through extensive train-time optimization. These methods fall into two main categories: supervised fine-tuning (SFT) with large-scale annotated datasets~\citep{qin2025uitarspioneeringautomatedgui,liu2025infiguiagentmultimodalgeneralistgui, wu2024osatlasfoundationactionmodel} and reinforcement learning with carefully designed reward functions~\citep{lu2025uir1enhancingefficientaction,liu2025infiguir1advancingmultimodalgui, luo2025guir1generalistr1style,tang2025guig2gaussianrewardmodeling}. However, these approaches share two fundamental limitations. First, they rely exclusively on train-time scaling while leaving test-time computation underutilized, missing opportunities for performance gains through inference-time optimization. Second, they require extensive labeled data, where each sample demands precise pixel-level annotations, creating a significant bottleneck for scaling to new domains and applications~\citep{chu2025sftmemorizesrlgeneralizes,guiactor}.

This raises a critical question: \textbf{\textit{{Can we leverage test-time computation to enhance GUI grounding performance without relying on any additional labeled data?}}} Recent breakthroughs in large language models have demonstrated the remarkable potential of test-time scaling~\citep{guan2025rstarmathsmallllmsmaster,snell2024scalingllmtesttimecompute,muennighoff2025s1simpletesttimescaling}. Self-consistency~\citep{wang2023selfconsistencyimproveschainthought} aggregates multiple reasoning paths through majority voting, achieving substantial improvements in mathematical reasoning. Test-time reinforcement learning (TTRL)~\citep{zuo2025ttrltesttimereinforcementlearning} enables models to self-improve on unlabeled data by generating experiences and computing rewards during inference. These successes in language domains suggest potential for applying test-time scaling to vision-language tasks like GUI grounding.

However, adapting test-time scaling to GUI grounding presents unique challenges. Unlike text-based reasoning where outputs are discrete tokens, GUI grounding operates in a continuous, high-resolution coordinate space where minor pixel deviations can lead to incorrect element selection~\citep{yang2025gta1guitesttimescaling}. The visual complexity of modern interfaces, with overlapping elements, dynamic layouts, and varying resolutions, introduces significant prediction uncertainty. The key insight is to transform this uncertainty from a limitation into an opportunity: when models generate multiple predictions for the same element, the patterns of agreement and disagreement across predictions reveal valuable information about localization confidence of the model.

In this work, we present GUI-RC (GUI Region Consistency), a test-time scaling approach that aggregates spatial information across multiple model predictions to improve grounding accuracy. Our core observation is that when sampling multiple predictions from a model, certain screen regions consistently appear across different outputs, indicating higher confidence in those locations. By constructing a spatial voting mechanism that identifies consensus regions with maximum overlap, GUI-RC achieves significant performance improvements (e.g., +2.75\% on OS-Atlas-Base-7B) without any additional training or labeled data.

Building upon this foundation, we introduce GUI-RCPO (GUI Region Consistency Policy Optimization), which enables test-time training through region consistency signals. Inspired by recent advances in TTRL~\citep{zuo2025ttrltesttimereinforcementlearning}, GUI-RCPO computes rewards based on how well each prediction aligns with the consensus across multiple samples, then uses these self-generated rewards to update model parameters during inference. 
This label-free optimization further improves performance using only 1,272 unlabeled data (e.g., +5.5\% on Qwen2.5-VL-3B-Instruct), demonstrating that models can effectively self-improve on unlabeled data.

Our main contributions are:

\begin{itemize}
 \item We propose GUI-RC, a test-time scaling method for GUI grounding that leverages spatial voting across multiple predictions to improve localization accuracy without additional training or labeled data.
 \item We introduce GUI-RCPO, a test-time reinforcement learning method that uses region consistency as a self-supervised reward signal, enabling models to improve grounding capabilities through policy optimization on unlabeled GUI screenshots.
 \item We demonstrate consistent improvements across multiple benchmarks and model architectures. GUI-RC improves accuracy by 2-3\%, while GUI-RCPO achieves further gains of 3-6\% through label-free optimization.
 \item We reveal that further applying GUI-RC after GUI-RCPO yields additional performance gains, demonstrating that our methods support progressive, self-bootstrapping improvement without external supervision, and provide a complementary alternative to train-time optimization for GUI automation.
\end{itemize}

\section{Related Works}

\subsection{GUI Grounding}
GUI grounding refers to the task of locating target elements on a screen based on natural language instructions. Given a screenshot $s$ and an instruction $i$, the model $M$ is required to output the specific position of the target element. Current approaches primarily fall into two paradigms. The first formulates GUI grounding as point prediction, directly outputting the coordinate $(x,y)$. These include supervised fine-tuning methods~\citep{cheng2024seeclickharnessingguigrounding,lin2024showuivisionlanguageactionmodelgui,wu2024oscopilotgeneralistcomputeragents,xu2025aguvisunifiedpurevision} and reinforcement learning methods~\citep{shi2025mobileguirladvancingmobilegui,shi2025mobilegui,lu2025ui,liu2025infiguig1advancingguigrounding}. The second paradigm predicts bounding boxes $(x^-, y^-, x^+, y^+)$ representing the region that best matches the instruction. Representative works include OS-Atlas~\citep{wu2024osatlasfoundationactionmodel} for cross-platform corpus development, and reinforcement learning approaches~\citep{tang2025guig2gaussianrewardmodeling,zhou2025guig1understandingr1zeroliketraining} for region-based grounding optimization. While all existing methods rely heavily on train-time optimization with labeled data, our work takes an orthogonal approach by leveraging test-time computation to improve grounding accuracy without additional training.

\subsection{Test-Time Scaling}
Test-time strategies for LLMs have shown that increasing inference computation can substantially enhance output accuracy without altering model weights~\citep{wei2022chain, madaan2023self, liu2025rethinking}. Representative strategies include self-consistency voting that aggregates multiple outputs to select the most confident~\citep{wang2023selfconsistencyimproveschainthought}, tree-based methods exploring diverse reasoning paths~\citep{yao2023tree,muennighoff2025s1simpletesttimescaling}, and test-time reinforcement learning (TTRL) enabling models to iteratively refine their outputs using self-generated rewards~\citep{zuo2025ttrltesttimereinforcementlearning}. Extending these strategies to GUI grounding introduces challenges due to continuous nature of coordinate predictions. Existing test-time strategies for GUI grounding primarily rely on zoom-based refinement, such as methods proposed by \citet{wu2025dimoguiadvancingtesttimescaling} and \citet{luo2025visualtesttimescalinggui}. Our methods introduce a spatial voting mechanism that transforms the uncertainty in continuous predictions into reliable consensus regions, enabling effective test-time scaling that works across both point-based and region-based grounding paradigms, without requiring specialized preprocessing.

\section{Methods}

\begin{figure*}[t]
    \centering
    \includegraphics[width=1\linewidth]{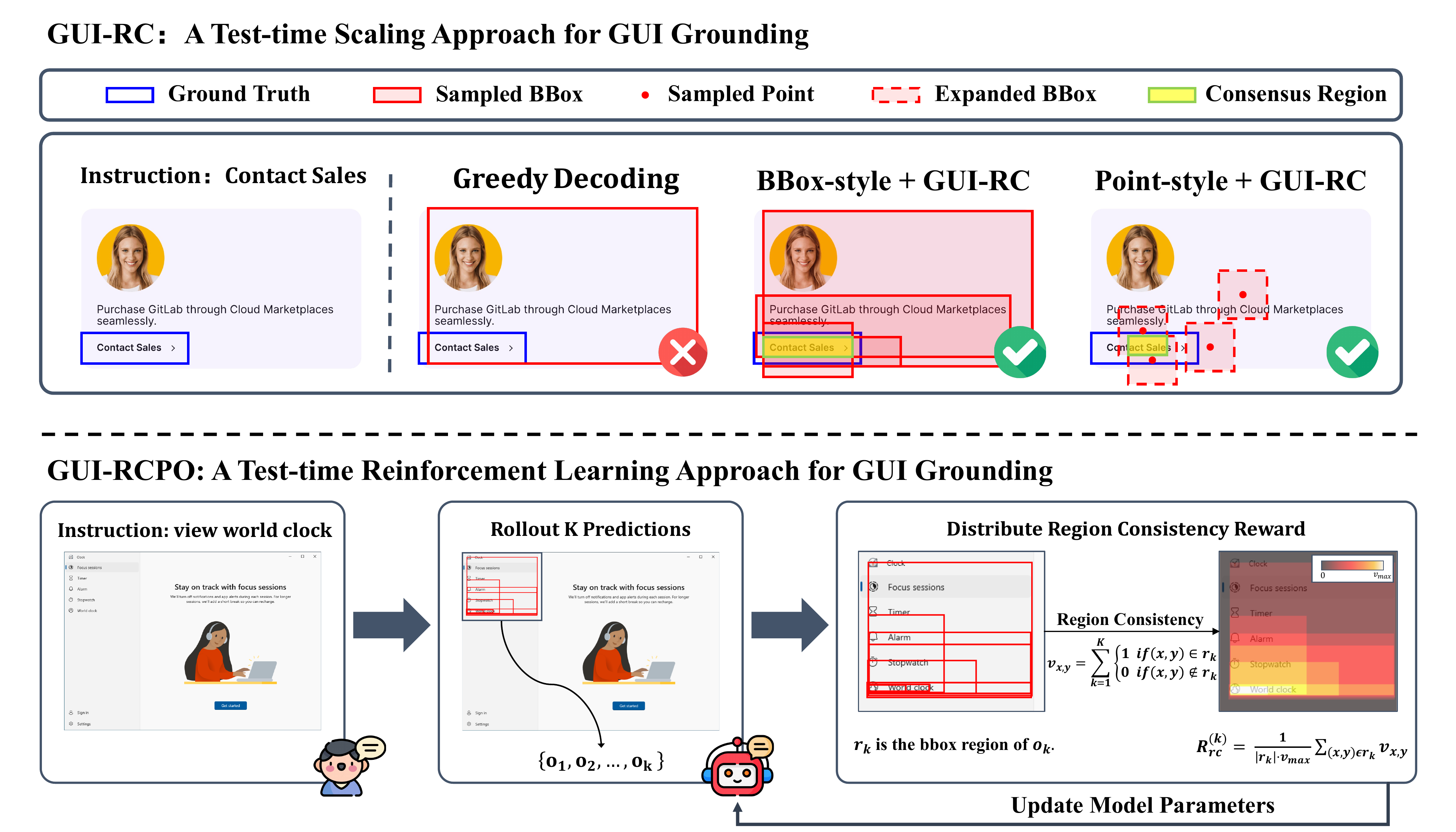}
    \caption{Overview of our test-time scaling methods for GUI grounding. \textbf{Upper}: GUI-RC aggregates $K$ sampled predictions through spatial voting to extract a consensus region, achieving more accurate localization than greedy decoding. \textbf{Lower}: GUI-RCPO computes region consistency rewards based on the voting heatmap and uses these self-supervised signals to update model parameters, enabling label-free improvement through test-time reinforcement learning.}
    \label{fig:gui_rcpo}
\end{figure*}

We first formalize the GUI grounding task and identify the key challenges in applying test-time scaling to this domain. We then present GUI-RC, our test-time scaling approach that leverages spatial consistency across multiple predictions to improve grounding accuracy. Finally, we introduce GUI-RCPO, which extends region consistency to reward signals for test-time reinforcement learning on unlabeled GUI data. Figure~\ref{fig:gui_rcpo} provides an overview of both methods.

\subsection{Problem Formulation}
\label{md:formulation}

GUI grounding aims to map natural language instructions to precise locations on graphical interfaces. Formally, given a screenshot $s \in \mathbb{R}^{H \times W \times 3}$ and an instruction $i$, a model $M$ outputs the spatial location of the UI element that best matches the instruction. As discussed in the previous section, this task has two primary formulations:
\begin{equation}
\text{Point-based: } M(s, i) \rightarrow (x, y)
\end{equation}
\begin{equation}
\text{Region-based: } M(s, i) \rightarrow (x^-, y^-, x^+, y^+)
\end{equation}
Both formulations face a fundamental challenge: the continuous nature of coordinate prediction introduces inherent uncertainty. This uncertainty is compounded by the complexity of modern interfaces with overlapping elements, varying resolutions, and ambiguous natural language descriptions.

Our key insight is that, rather than regarding this uncertainty as a limitation, it can be leveraged for improvement. When a model generates multiple predictions for the same input, their spatial distribution reveals implicit confidence patterns that can guide more accurate localization.

\label{md:formulation}

\subsection{GUI-RC: Test-Time Scaling via Spatial Voting}

\label{md:rc}

GUI-RC transforms the uncertainty in individual predictions into robust consensus through spatial aggregation. As illustrated in the upper part of Figure~\ref{fig:gui_rcpo}, while greedy decoding fails to locate the correct element, GUI-RC successfully identifies the target by aggregating information across multiple samples. The method operates in three stages: multi-sample generation, spatial voting, and consensus extraction.

\textbf{Multi-Sample Generation.} Given an input pair $(s, i)$, we sample $K$ predictions from the model using temperature-based sampling:
\begin{equation}
\{M_k(s, i)\}_{k=1}^K \sim p_\theta(\cdot | s, i)
\end{equation}
This sampling process naturally produces variation in predictions due to the continuous output space and model uncertainty. Crucially, we observe that despite this variation, certain screen regions appear consistently across samples, indicating higher model confidence.

\textbf{Spatial Voting Mechanism.} To quantify this consistency, we construct a spatial voting grid $v \in \mathbb{R}^{H \times W}$ matching the screenshot resolution. Each sampled prediction contributes votes to this grid:
\begin{equation}
v_{x,y} = \sum_{k=1}^{K} \mathds{1}[(x,y) \in R_k]
\end{equation}
where $R_k$ represents the region from the $k$-th prediction. For region-based models, $R_k$ is simply the predicted bounding box. For point-based models, we approximate the implicit attention region by expanding the predicted point $(x_k, y_k)$ into a square region of size \begin{equation}
R_k = [x_k - \alpha/2, x_k + \alpha/2] \times [y_k - \alpha/2, y_k + \alpha/2]
\end{equation}
This expansion is necessary because point predictions, while precise, do not explicitly encode the spatial extent of the target element. As shown in Figure~\ref{fig:gui_rcpo}, the expanded regions for point-style models (dashed red boxes) enable meaningful spatial voting similar to bbox-style models.

\textbf{Consensus Extraction.} The voting grid $v$ now encodes the collective attention of the model across samples. We extract the consensus region through a principled selection process that first identifies the maximum vote count across the entire grid as $v_{\max} = \max_{x,y} v_{x,y}$. This maximum value represents the highest level of agreement among predictions. We then find all contiguous regions where every pixel has this maximum vote count, forming the set $\mathcal{R}_{v_{\max}} = \{r : \forall (x,y) \in r, v_{x,y} = v_{\max}\}$. Among these high-confidence regions, we select the one with the largest area as our final consensus region: $\hat{r}_{\text{cons}} = \arg\max_{r \in \mathcal{R}_{v_{\max}}} |r|$. The consensus region $\hat{r}_{\text{cons}}$ represents the area where the model shows highest and most consistent attention, providing a more reliable grounding prediction than any individual sample.

\subsection{GUI-RCPO: Test-Time Reinforcement Learning via Region Consistency}
\label{md:rcpo}

While GUI-RC improves performance through inference-time aggregation, we further explore whether region consistency can guide model improvement through test-time training. As illustrated in the lower part of Figure~\ref{fig:gui_rcpo}, GUI-RCPO transforms region consistency into a self-supervised reward signal for policy optimization.

\textbf{Region Consistency as Reward.} The key insight is that predictions aligning with high-consistency regions should be reinforced, while outliers should be suppressed. For each sampled prediction $r_k$ in the rollout, we compute its region consistency reward:
\begin{equation}
R_{rc}^{(k)} = \frac{1}{|r_k| \cdot v_{\max}} \sum_{(x,y) \in r_k} v_{x,y}
\end{equation}
This reward measures the average vote density within the predicted region, normalized by the region size and maximum possible votes. As visualized in Figure~\ref{fig:gui_rcpo}, the heatmap representation shows how different regions receive varying levels of votes, with warmer colors indicating higher consistency. Predictions that overlap with these high-vote regions receive higher rewards, encouraging the model to converge toward consensus areas.

\textbf{Policy Optimization.} We formulate GUI grounding as a reinforcement learning problem where the VLM acts as policy $\pi_\theta$. Using Group Relative Policy Optimization (GRPO)~\citep{shao2024deepseekmathpushinglimitsmathematical}, we optimize the expected region consistency reward:
\begin{equation}
\mathcal{L}(\theta) = -\mathbb{E}_{(s,i) \sim D} \mathbb{E}_{r \sim \pi_\theta(\cdot|s,i)} [A(r) \log \pi_\theta(r|s,i)]
\end{equation}
where $A(r)$ is the advantage computed from relative rewards within each group of samples. GRPO's group-relative formulation is particularly suitable for our setting as it normalizes rewards across different inputs, preventing optimization bias toward easier examples.

A unique property of GUI-RCPO is its ability to progressively improve without external supervision. As the model updates its parameters based on region consistency rewards, its predictions become more concentrated around high-confidence regions, which in turn provides stronger and more reliable reward signals for further optimization. This self-bootstrapping process continues until the model converges to a stable distribution centered on consensus regions.

\section{Experiments}

\subsection{Experiment Setup}

\begin{table*}[t!]
\centering
\setlength{\tabcolsep}{5pt}
\resizebox{1.0\linewidth}{!}{
\begin{tabular}{l lllllllll}
\toprule 
& \multicolumn{2}{c}{\textbf{SSv2.Mobile}} & \multicolumn{2}{c}{\textbf{SSv2.Desktop}} & \multicolumn{2}{c}{\textbf{SSv2.Web}} &  \multirow{2}{*}{\textbf{SSv2.avg}} & \multirow{2}{*}{\textbf{SSv1.avg}} & \multirow{2}{*}{\textbf{SSPro.avg}}\\ 
\cmidrule(lr){2-3} \cmidrule(lr){4-5} \cmidrule(lr){6-7}
& \multicolumn{1}{c}{Text} & \multicolumn{1}{c}{Icon} & \multicolumn{1}{c}{Text} & \multicolumn{1}{c}{Icon} & \multicolumn{1}{c}{Text} & \multicolumn{1}{c}{Icon} & & \\
\midrule 
\multicolumn{10}{l}{\textit{General Models}} \\ 
\midrule 
\pointstyle InternVL3-2B-Instruct & 89.92 & 76.44 & 38.89 & 26.19 & 46.43 & 25.32 & 52.75 & 51.02 & 1.03 \\
\pointstyle w/ \textbf{GUI-RC} & 89.92 & 77.49↑ & 38.33 & 24.60 & 46.07 & 27.00↑ & 52.91\scriptsize{+0.16} & 52.20\scriptsize{+1.18} & 1.33\scriptsize{+0.30} \\
\midrule 
\pointstyle InternVL3-8B-Instruct & 94.19 & 79.58 & 79.44 & 53.17 & 91.07 & 71.73 & 80.97 & 79.72 & 13.28 \\
\pointstyle w/ \textbf{GUI-RC} & 94.19 & 81.15↑ & 80.56↑ & 56.35↑ & 91.07 & 71.73 & 81.68\scriptsize{+0.71} & 80.03\scriptsize{+0.31} & 12.46\scriptsize{-0.82} \\
\midrule 
\bboxstyle Qwen2.5-VL-3B-Instruct & 97.67 & 75.92 & 85.56 & 59.52 & 84.64 & 65.82 & 80.11 & 76.97 & 20.18 \\
\bboxstyle w/ \textbf{GUI-RC} & 98.84↑ & 77.49↑ & 90.00↑ & 64.29↑ & 87.14↑ & 67.93↑ & 82.63\scriptsize{+2.52} & 78.46\scriptsize{+1.49} & 23.59\scriptsize{+3.41} \\
\midrule 
\bboxstyle Qwen2.5-VL-7B-Instruct & 98.84 & 84.29 & 86.67 & 73.81 & 88.57 & 78.90 & 86.48 & 84.20 & 19.80 \\
\bboxstyle w/ \textbf{GUI-RC} & 99.92↑ & 85.86↑ & 91.11↑ & 73.02 & 91.79↑ & 81.43↑ & 88.52\scriptsize{+2.04} & 85.53\scriptsize{+1.33} & 23.97\scriptsize{+4.17} \\
\midrule 
\multicolumn{10}{l}{\textit{GUI-specific Models}} \\ 
\midrule 
\pointstyle UGround-V1-7B & 96.51 & 82.72 & 96.11 & 82.54 & 92.50 & 83.12 & 89.62 & 87.11 & 31.50 \\
\pointstyle w/ \textbf{GUI-RC} & 96.51 & 83.77↑ & 95.56 & 84.13↑ & 92.86↑ & 81.43 & 89.62\scriptsize{+0.00} & 87.34\scriptsize{+0.23} & 31.63\scriptsize{+0.13} \\
\midrule 
\pointstyle UI-TARS-1.5-7B & 96.51 & 86.39 & 95.00 & 87.30 & 88.21 & 86.50 & 90.17 & 87.74 & 40.92 \\
\pointstyle w/ \textbf{GUI-RC} & 96.12 & 86.91↑ & 96.11↑ & 90.48↑ & 90.36↑ & 86.50 & 91.12\scriptsize{+0.95} & 88.52\scriptsize{+0.78} & 41.18\scriptsize{+0.26} \\
\midrule 
\bboxstyle OS-Atlas-Base-7B & 91.47 & 72.25 & 88.33 & 64.29 & 86.43 & 72.57 & 80.82 & 79.80 & 18.41 \\
\bboxstyle w/ \textbf{GUI-RC} & 91.47 & 78.53↑ & 88.89↑ & 68.25↑ & 89.29↑ & 76.37↑ & 83.57\scriptsize{+2.75} & 81.45\scriptsize{+1.65} & 19.67\scriptsize{+0.16} \\
\midrule 
\bboxstyle GUI-G2-7B & 99.61 & 92.15 & 95.56 & 88.89 & 95.00 & 88.61 & 93.79 & 91.51 & 46.43 \\
\bboxstyle w/ \textbf{GUI-RC} & 99.61 & 93.19↑ & 95.56 & 88.1 & 95.00 & 88.61 & 93.87\scriptsize{+0.08} & 91.90\scriptsize{+0.39} & 47.88\scriptsize{+1.45} \\
\bottomrule
\end{tabular}
}
\caption{Performance (\%) of the proposed test-time scaling method \textbf{GUI-RC} across GUI-Grounding benchmarks. Icons refer to output styles: Point (\pointstyle), Bounding-box (\bboxstyle).}
\label{tab:main_tts_results_with_gui_rc}
\end{table*}

\begin{table*}[t!]
\centering
\setlength{\tabcolsep}{5pt}
\resizebox{1.0\linewidth}{!}{
\begin{tabular}{l l l l l l l l l l}
\toprule 
& \multicolumn{2}{c}{\textbf{SSv2.Mobile}} & \multicolumn{2}{c}{\textbf{SSv2.Desktop}} & \multicolumn{2}{c}{\textbf{SSv2.Web}} &  \multirow{2}{*}{\textbf{SSv2.avg}} & \multirow{2}{*}{\textbf{SSv1.avg}} & \multirow{2}{*}{\textbf{SSPro.avg}}\\ 
\cmidrule(lr){2-3} \cmidrule(lr){4-5} \cmidrule(lr){6-7}
& \multicolumn{1}{c}{Text} & \multicolumn{1}{c}{Icon} & \multicolumn{1}{c}{Text} & \multicolumn{1}{c}{Icon} & \multicolumn{1}{c}{Text} & \multicolumn{1}{c}{Icon} & & \\
\midrule 
\multicolumn{10}{l}{\textit{General Models}} \\ 
\midrule 

\bboxstyle Qwen2.5-VL-3B-Instruct & 97.67 & 75.92 & 85.56 & 59.52 & 84.64 & 65.82 & 80.11 & 76.97 & 20.18 \\
\bboxstyle w/ \textbf{GUI-RCPO} & 98.06↑ & 81.68↑ & 91.11↑ & 65.08↑ & 90.71↑ & 73.42↑ & 85.14\scriptsize{+5.03} & 82.47\scriptsize{+5.50} & 24.67\scriptsize{+4.49} \\
\midrule 

\bboxstyle Qwen2.5-VL-7B-Instruct & 98.84 & 84.29 & 86.67 & 73.81 & 88.57 & 78.90 & 86.48 & 84.20 & 19.80 \\
\bboxstyle w/ \textbf{GUI-RCPO} & 98.84 & 87.43↑ & 91.11↑ & 76.19↑ & 92.5↑ & 80.17↑ & 88.92\scriptsize{+2.48} & 86.64\scriptsize{+2.44} & 25.93\scriptsize{+6.13} \\

\midrule 
\multicolumn{10}{l}{\textit{GUI-specific Models}} \\ 
\midrule 
\pointstyle UI-TARS-1.5-7B & 96.51 & 86.39 & 95.00 & 87.30 & 88.21 & 86.50 & 90.17 & 87.74 & 40.92 \\
\pointstyle w/ \textbf{GUI-RCPO} & 97.29↑ & 86.39 & 97.22↑ & 82.54 & 91.07↑ & 87.34↑ & 90.96\scriptsize{+0.79} & 88.60\scriptsize{+0.86} & 41.43\scriptsize{+0.51} \\
\bottomrule

\end{tabular}
}
\caption{Performance (\%) of the proposed test-time reinforcement learning method \textbf{GUI-RCPO} across GUI-Grounding benchmarks. Icons refer to output styles: Point (\pointstyle), Bounding-box (\bboxstyle).}
\label{tab:main_ttrl_results_with_gui_rc}
\end{table*}

\textbf{Models.} We evaluate our methods on a diverse VLMs to demonstrate their generality across different architectures and training paradigms. For general-purpose models, we use Qwen2.5-VL-3B-Instruct and Qwen2.5-VL-7B-Instruct~\citep{bai2025qwen25vltechnicalreport}, as well as InternVL3-2B-Instruct and InternVL3-8B-Instruct~\citep{zhu2025internvl3exploringadvancedtraining}, which represent state-of-the-art vision-language models at different scales. For GUI-specific models that have been explicitly trained on GUI grounding tasks, we evaluate UGround-V1-7B~\citep{gou2024navigatingdigitalworldhumans}, OS-Atlas-Base-7B~\citep{wu2024osatlasfoundationactionmodel}, UI-TARS-1.5-7B~\citep{ui-tars-15-seed}, and GUI-G2-7B~\citep{tang2025guig2gaussianrewardmodeling}. These models span both point-based and region-based prediction paradigms, allowing us to assess the effectiveness of our methods across different output formats.

\paragraph{Evaluation Benchmarks and Metrics.} We evaluate our methods on three GUI grounding benchmarks: ScreenSpot~\citep{cheng2024seeclickharnessingguigrounding}, ScreenSpot-v2~\citep{wu2024osatlasfoundationactionmodel}, and ScreenSpot-Pro~\citep{li2024screenspot-pro}. ScreenSpot and ScreenSpot-v2 assess model's grounding performance in general GUI environments spanning Mobile, Web, and Desktop platforms. ScreenSpot-Pro specifically focuses on high-resolution and professional interfaces. 
Following standard evaluation protocols, we adopt grounding accuracy as our primary metric: a prediction is considered correct if the predicted point or the center of the predicted bounding box falls in the ground-truth bounding box~\citep{cheng2024seeclickharnessingguigrounding}.

\paragraph{Implementation Details.} For GUI-RC, we sample 64 outputs using a temperature of 0.5 and a top\_p of 0.95 for voting, the hyperparameter \(\alpha\) is set to 50. For the baselines, we employ greedy decoding with temperature 0.
For GUI-RCPO, we adopt the VLM-R1~\citep{shen2025vlmr1stablegeneralizabler1style} framework and conduct TTRL training on the Screenspot-v2 benchmark without using the ground-truth data. For each input, 16 samples are generated with a temperature of 0.7 and top\_p of 0.95. We train the models for 2 epochs (approx. 40 steps) with a global batch size of 64, learning rate of 1e-6, and KL penalty $\beta = 0.04$. All training and evaluation are conducted on 8 NVIDIA A100-80GB GPUs.

\subsection{Main Results}
\subsubsection{Evaluation Results of GUI-RC Experiments}

\paragraph{GUI-RC consistently improves the end-to-end grounding  performance.} We compared the performance of the base models on three benchmarks before and after applying GUI-RC. Table~\ref{tab:main_tts_results_with_gui_rc} presents the evaluation results.
It can be observed that GUI-RC consistently improves the overall grounding capability across different models, regardless of its output style and whether the model is specifically trained for GUI tasks. 
For instance, OS-Atlas-Base-7B achieves an overall improvement of 2.75\%, with a notable 6.28\% increase in icon localization in mobile scenarios. 
Moreover, for general models like Qwen2.5-VL-3B/7B-Instruct that output in bbox-style, GUI-RC brings even greater improvements on the ScreenSpot-Pro compared to ScreenSpot and ScreenSpot-v2. This suggests that GUI-RC is particularly effective in helping models tackle more challenging grounding tasks involving high-resolution and professional GUIs.

\paragraph{GUI-RC achieves greater improvements when applied to bbox-style prediction models.} Another observation is that GUI-RC provides greater improvements for models that output bounding boxes compared to those output points. This is because when models predict bounding boxes, the bounding boxes inherently reflect the regions that the models are attending to. In contrast, for models predict points, when we manually expand a point into a bounding box, we are simulating the model's attention region. This may fail to accurately represent the actual region that the model focuses on, which further introduces biases in identifying the consensus regions, thus limiting the performance of GUI-RC. Nevertheless, GUI-RC still brings improvements to most point-style prediction models, indicating its robustness.

\subsubsection{Evaluation Results of GUI-RCPO Experiments}

\paragraph{GUI-RCPO is supervised by GUI-RC yet outperforms it.}
We compare the performance of base models with and without further TTRL training via GUI-RCPO on the three benchmarks. As Table~\ref{fig:gui_rcpo} shows, GUI-RCPO also brings consistent improvements and even outperforms GUI-RC. For instance, GUI-RC brings an improvement of 1.49\% for Qwen2.5-VL-3B-Instruct on ScreenSpot, while after GUI-RCPO training, it achieves an impressive gain of 5.5\%. Intuitively, the performance upper bound of GUI-RCPO should be that of GUI-RC, as it utilizes the region consistency as a reward signal for RL. However, in practice, GUI-RCPO not only matches but exceeds GUI-RC, which aligns with prior findings in TTRL~\citep{zuo2025ttrltesttimereinforcementlearning}. This indicates that the model practically learns a more effective GUI grounding strategy through GUI-RCPO, rather than merely fitting to the consensus region. 
Notably, GUI-RCPO further brings performance gains for models that have already been specifically trained on GUI tasks, indicating the effectiveness of introducing the region consistency reward. 

\paragraph{GUI-RCPO generalizes well in out-of-distribution scenarios.}
Although models are trained on Screenspot-v2, GUI-RCPO also shows significant improvement on ScreenSpot-Pro, which is an out-of-distribution benchmark featuring high-resolution and domain-specific GUIs. This further proves that GUI-RCPO does not rely on overfitting but genuinely enhances the model's general GUI grounding capability. Unlike direct fine-tuning on labeled data, which risks overfitting to the resolution and layout of the training set, GUI-RCPO enables robust generalization across different screen resolutions and interface layouts. 

\section{Analysis}

\subsection{Ablation Studies on Decoding Strategy of GUI-RC}
\begin{figure*}[t]
    \centering
    \includegraphics[width=1\linewidth]{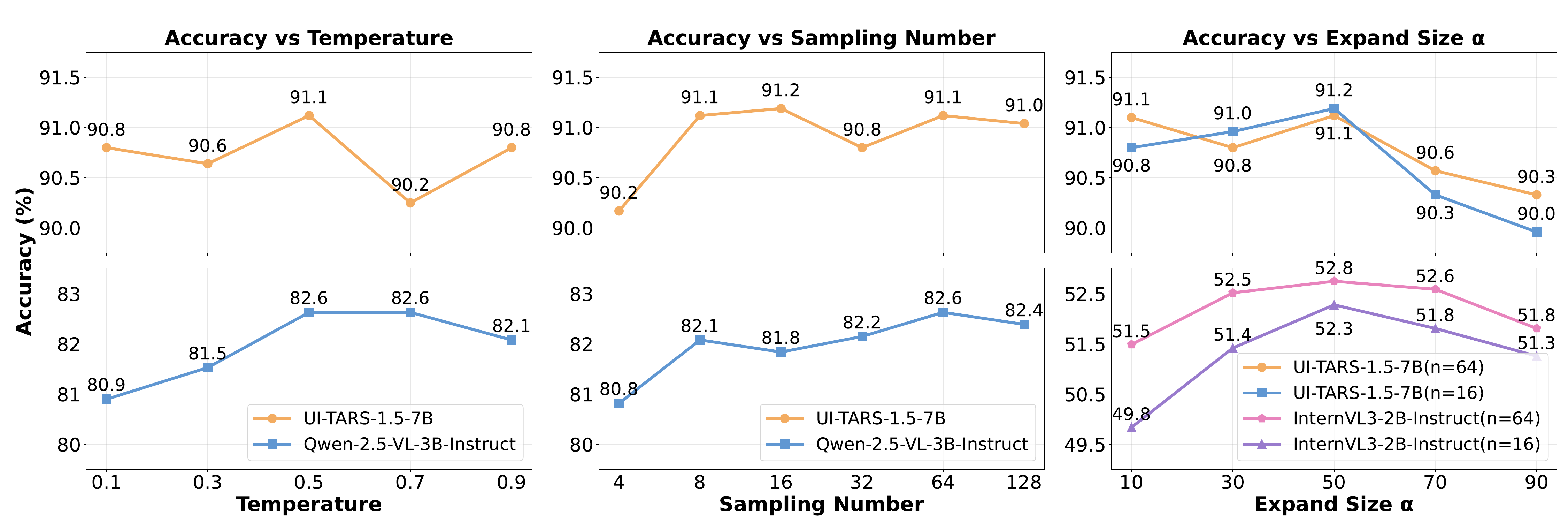}
    \caption{Ablation study results on ScreenSpot-v2 with varying temperature, sampling number, and hyperparameter $\alpha$.}
    \label{fig:ablation}
\end{figure*}

We conduct ablation studies on the decoding strategy of GUI-RC to analyze how different parameters affect its GUI grounding performance. Specifically, we first fix the temperature, sampling number, and expand size hyperparameter $\alpha$ to 0.5, 64, and 50 respectively, then vary each parameter individually to observe its impact on GUI-RC. We employ Qwen2.5-VL-3B-Instruct (representing bbox-style prediction models), UI-TARS-1.5-7B and InternVL3-2B-Instruct (representing point-style prediction models) for ablation studies on the ScreenSpot-v2 benchmark. The results are shown in Figure~\ref{fig:ablation}.

\paragraph{Temperature.}
The performance of GUI-RC generally exhibits an increasing-then-decreasing trend as the temperature rises. As temperature controls the diversity of sampled outputs during decoding, a high temperature encourages broad exploration but also increases instability in model generation. Therefore, a moderate increase in temperature helps the model explore broader regions while generating relatively concentrated outputs. However, further increasing the temperature would cause the predicted regions to become overly dispersed, making it difficult to obtain a focused and reliable consensus region.

\paragraph{Sampling Number.}
As the number of sampled predictions increases, the performance of GUI-RC initially improves and then gradually plateaus. This is because with more samples, the distribution of predicted regions becomes more stable, leading the consensus region to converge toward a fixed area. Once the predictions reach sufficient diversity and coverage, additional samples contribute diminishing improvements, leading to performance saturation.

\begin{figure*}[t!]
    \centering
    \includegraphics[width=1\linewidth]{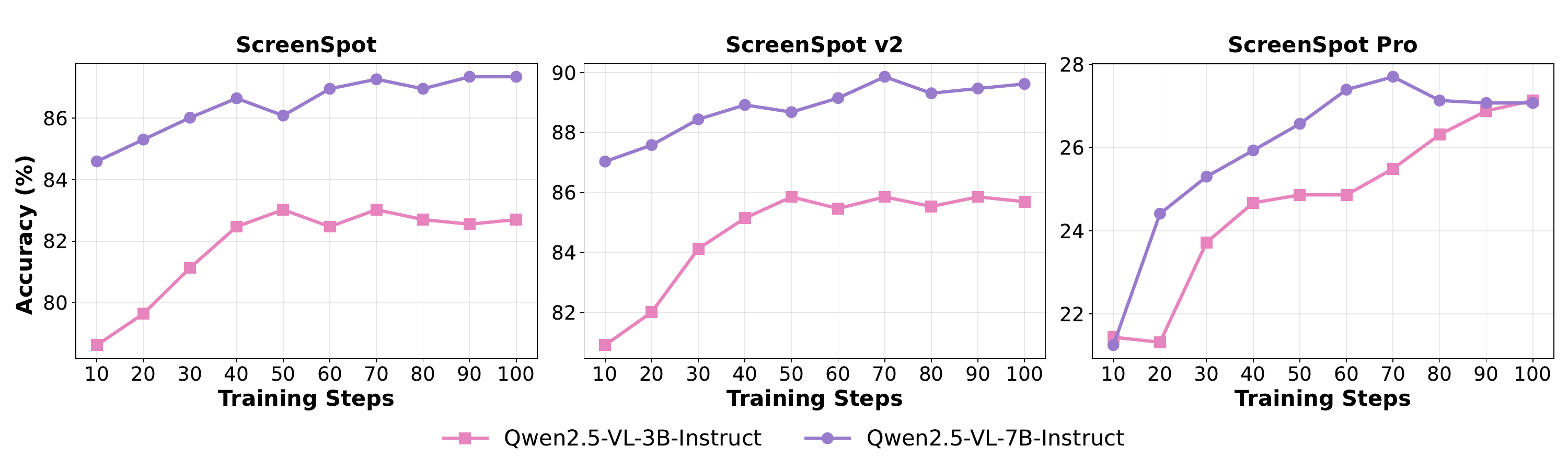}
    \caption{Accuracy (\%) across training steps of Qwen2.5-VL-3B-Instruct and Qwen2.5-VL-7B-Instruct throughout GUI-RCPO.}
    \label{fig:acc_step}
\end{figure*}

\paragraph{Hyperparameter \(\alpha\).}
The performance of GUI-RC follows an increasing-then-decreasing trend as \(\alpha\) grows. As the hyperparameter \(\alpha\) primarily affects the estimation of the attention area for point-style prediction models, both overly small and large expansion sizes introduce deviations in estimating the area models actually attend to. This results in larger deviations in the computed consensus region, thereby impairing the grounding accuracy.

\subsection{GUI-RCPO Enables Consistent Improvements during Test-time Training}
We observe the performance trajectories of Qwen2.5-VL-3B-Instruct and Qwen2.5-VL-7B-Instruct during GUI-RCPO training on three benchmarks, as shown in Figure~\ref{fig:acc_step}. As training steps increase, the models' accuracy stably improves across all three GUI-Grounding benchmarks and converges around 80 steps. In particular, although the models are trained solely on ScreenSpot-v2, they do not exhibit overfitting to the training data or degradation on other benchmarks like ScreenSpot-Pro, demonstrating the robust generalization of GUI-RCPO.

\subsection{Applying GUI-RC after GUI-RCPO Leads to Further Improvements}

\begin{table}[h]
\centering
\resizebox{1.0\linewidth}{!}{
\begin{tabular}{l l l}
\toprule
\textbf{Model} & \textbf{SSv2.avg} & \textbf{SSPro.avg} \\
\midrule
\multicolumn{3}{l}{\textit{General Models}} \\
\midrule
\bboxstyle Qwen2.5-VL-3B-Instruct+GUI-RCPO & 85.14 & 24.67 \\
\bboxstyle w/ \textbf{GUI-RC} & 86.32\scriptsize{+1.18} & 26.19\scriptsize{+1.52}\\
\midrule
\bboxstyle Qwen2.5-VL-7B-Instruct+GUI-RCPO & 88.92 & 25.93 \\
\bboxstyle w/ \textbf{GUI-RC} & 89.78\scriptsize{+0.86} & 26.69\scriptsize{+0.76} \\
\bottomrule
\end{tabular}
}
\caption{Performance of applying GUI-RC to bbox-style prediction models after GUI-RCPO on ScreenSpot-v2 and ScreenSpot-Pro benchmarks. In this table, GUI-RC is performed with temperature = 1.0.}
\label{tab:GUI-RCPO+GUI-RC}
\end{table}

We further apply GUI-RC to the bbox-style prediction models after being trained with GUI-RCPO, and evaluate their performance on ScreenSpot-v2 and ScreenSpot-Pro. It is important to note that for GUI-RC in this experiment, we keep all other parameters consistent with previous settings, except increasing the sampling temperature to 1.0. This is because the reward signals in GUI-RCPO training encourage the model to predict more concentrated regions. Therefore, during GUI-RC, a higher decoding temperature is needed to encourage the model to explore broader regions. 

The evaluation results are shown in Table~\ref{tab:GUI-RCPO+GUI-RC}. It can be observed that even after GUI-RCPO training, applying GUI-RC voting mechanism still leads to additional performance gains. For instance, Qwen2.5-VL-3B-Instruct can gain an additional 1.52\% performance on ScreenSpot-Pro through GUI-RC even after GUI-RCPO. This indicates that our methods enable the models to recursively improve themselves without relying on any external supervision.

\section{Limitations}
\paragraph{Performance gain limited in point-style grounding.} Despite its effectiveness, GUI-RC has several limitations. As analyzed in the experiment section, GUI-RC brings relatively limited improvements for models with point-style outputs. Although most existing GUI Agents adopt point-style grounding in order to seamlessly integrate with subsequent action execution, an increasing number of recent studies~\citep{guiactor, zhou2025guig1understandingr1zeroliketraining} have pointed out the limitations of point-based prediction and the advantages of region-level supervision. Therefore, we expect that the strengths of GUI-RC will be amplified in future research.

\paragraph{Rely on the model's inherent capabilities.} Moreover, GUI-RC primarily addresses misleading and biased hallucinations in grounding, but it is hard to resolve confusion hallucinations (i.e., the predicted region fails to match any valid UI element). In other words, GUI-RC assumes that the model has a certain ability to recognize the target element. It can tolerate the model's predictions to be imprecise or biased, but not completely random or unrelated. Therefore, GUI-RC requires the model to be familiar with the GUI environment, but it does not require the model to be specifically trained on GUI tasks.

\section{Conclusion}

We introduce GUI-RC, a test-time scaling approach for GUI grounding that leverages region consistency across multiple predictions to enhance model performance without requiring additional training. Building on this idea, we further proposed GUI-RCPO, a test-time reinforcement learning method that transforms region consistency into a self-supervised reward signal, enabling models to self-improve during inference without the need for labeled data. Extensive experiments across a wide range of general and GUI-specific models demonstrate that our methods consistently improve GUI grounding performance and generalize well to out-of-distribution scenarios. Our findings reveal the untapped potential of test-time training for GUI agents and suggest a promising direction toward more robust and data-efficient GUI automation systems.

\section{Acknowledgments}

This work is supported by the National Natural Science Foundation of China (No. 62506332), National Key Research and Development Project (No. 2024YFB3312900), the Key Research and Development Program of Zhejiang Province, China (No. 2024C01034), the Fundamental Research Funds for the Central Universities (226-2024-00170), MOE Engineering Research Center of Digital Library, CIPS-LMG Huawei Innovation Fund and ZJU Kunpeng\&Ascend Center of Excellence.

\bibliography{aaai2026}

\clearpage
\section{Appendix}
\subsection{A. Pseudo-code of Region Consistency Reward Function}
\begin{lstlisting}[language=Python, caption={The pseudo-code of the region consistency reward function.}, label={lst:region_consistency_reward}]
def region_consistency_reward_fn(outputs, image_size):
    """
    Assigns a soft reward to each output based on how much the predicted bounding box consists with the most voted region.
    """
    # Initialize voting grid
    W, H = image_size
    grid = zeros(H, W)
    sampled_bboxes = []

    # Vote on each bounding box
    for output in outputs:
        bbox = extract_answer(output["content"])
        x1, y1, x2, y2 = int(bbox[0]), int(bbox[1]), int(bbox[2]), int(bbox[3])
        sampled_bboxes.append([x1, y1, x2, y2])
        grid[y1:y2, x1:x2] += 1

    # Find the most voted region
    max_vote = max(grid)

    # Compute reward for each bounding box
    rewards = []
    for bbox in sampled_bboxes:
        x1, y1, x2, y2 = bbox
        region = grid[y1:y2, x1:x2]
        region_sum = sum(region)
        region_area = (x2 - x1) * (y2 - y1)
        if region_area == 0:
            reward = 0
        else:
            reward = region_sum / (max_vote * region_area)
        rewards.append(reward)

    return rewards

def extract_answer(content, alpha):
    """
    Extracts bounding box coordinates from a text string.
    If 4 numbers are found, treat as a full box [x1, y1, x2, y2].
    If 2 numbers are found, treat as a single point [x, y] and convert to an alpha*alpha box.
    Otherwise, return a zero box.
    """
    numbers = find_all_numbers(content)
    
    if length(numbers) == 4:
        return [float(numbers[0]), float(numbers[1]),
                float(numbers[2]), float(numbers[3])]
    
    elif length(numbers) == 2:
        x, y = float(numbers[0]), float(numbers[1])
        return [x - alpha/2, y - alpha/2, x + alpha/2, y + alpha/2]
    
    else:
        return [0, 0, 0, 0]
\end{lstlisting}

\subsection{B. Case Studies on How GUI-RC Mitigates Hallucinations in GUI Grounding}
\begin{figure*}[t!]
    \centering

    \begin{subfigure}[t]{0.48\linewidth}
        \centering
        \includegraphics[width=\linewidth]{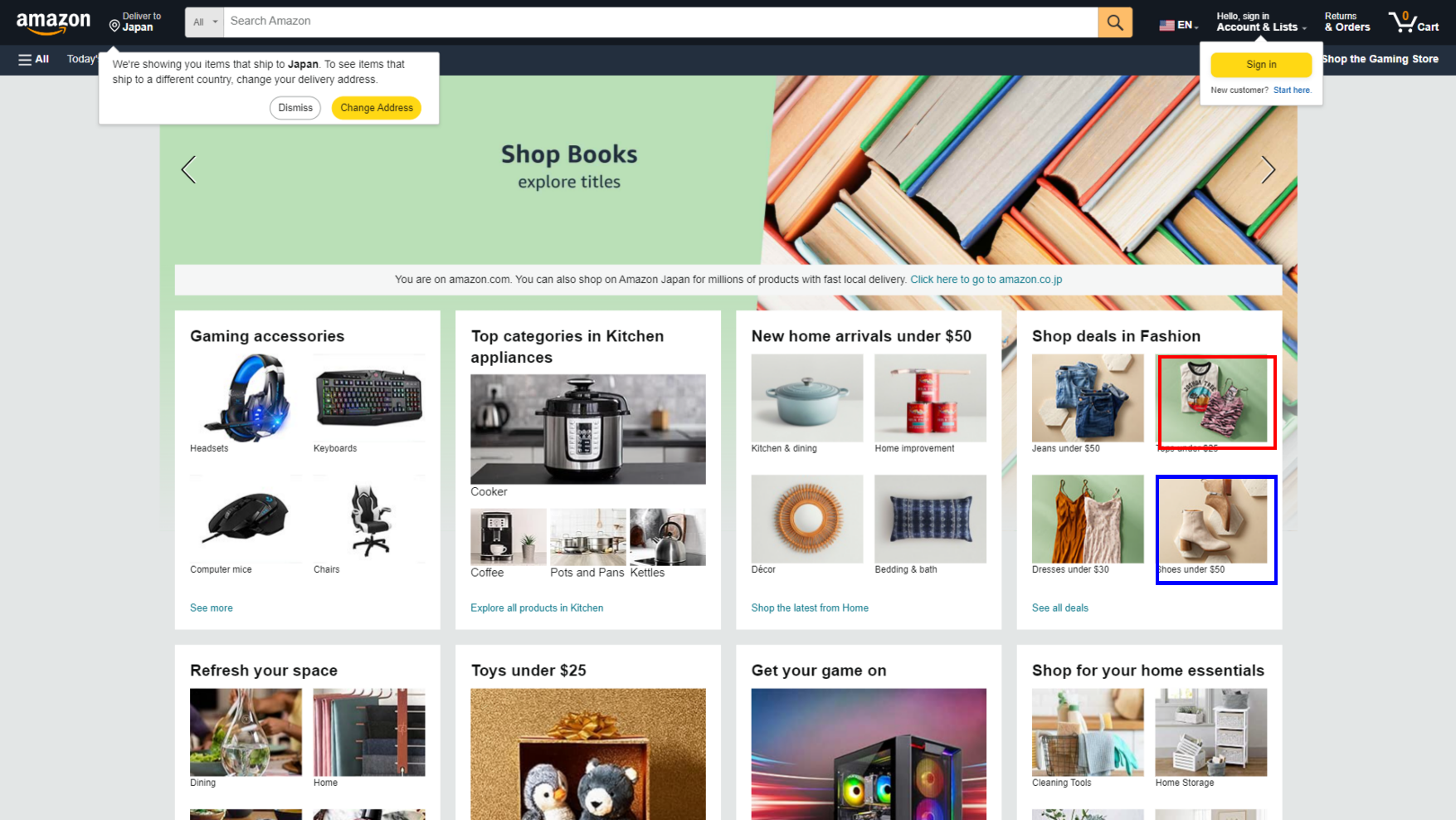}
        \caption{ScreenSpot-v2: ``check shoes under 50 dollars in `shop deals in fashion' part'' (Greedy)}
        \label{fig:rc_case1a}
    \end{subfigure}
    \hfill
    \begin{subfigure}[t]{0.48\linewidth}
        \centering
        \includegraphics[width=\linewidth]{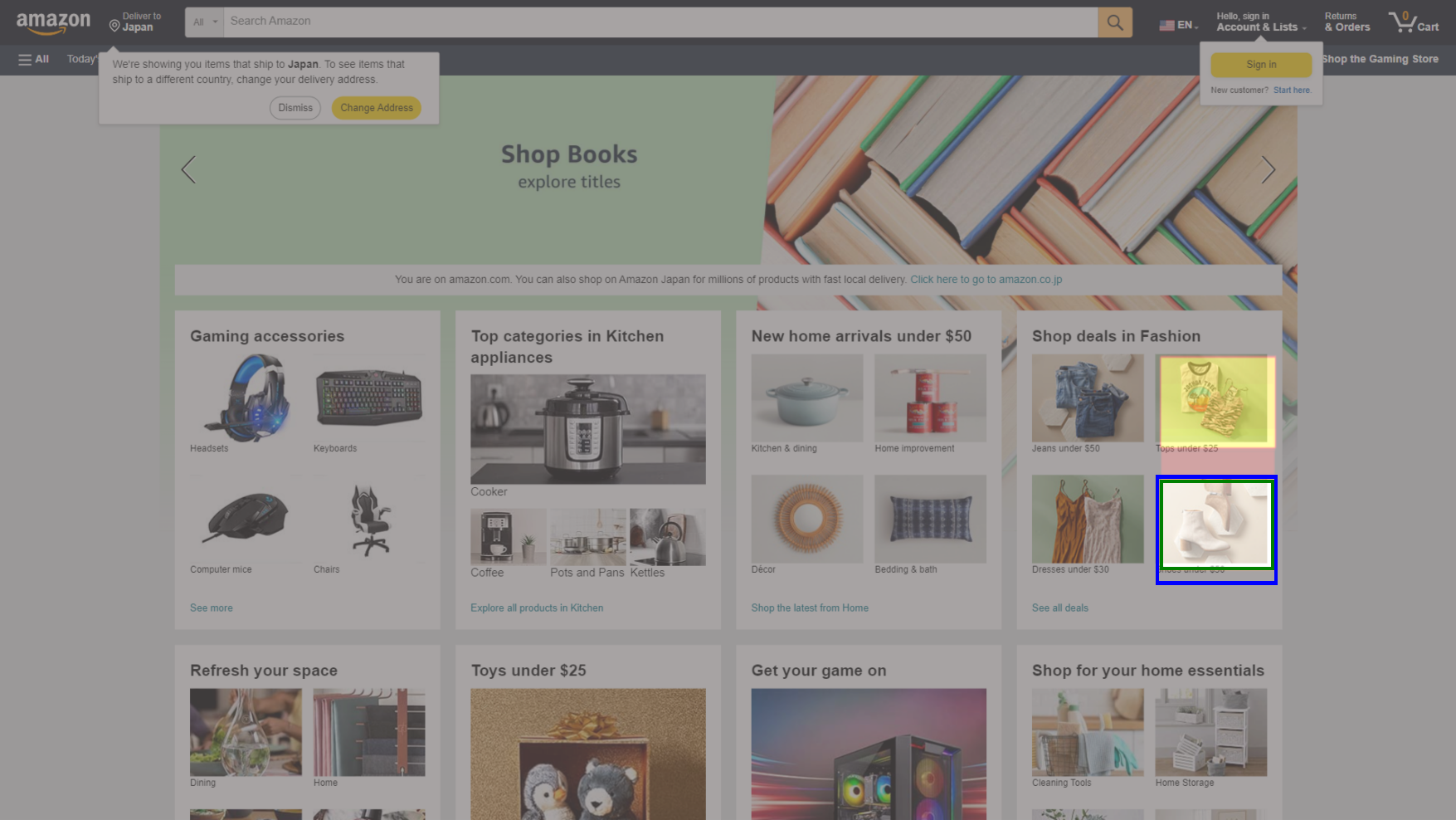}
        \caption{ScreenSpot-v2: ``check shoes under 50 dollars in `shop deals in fashion' part'' (GUI-RC)}
        \label{fig:rc_case1b}
    \end{subfigure}

    \vspace{1em}

    \begin{subfigure}[t]{0.48\linewidth}
        \centering
        \includegraphics[width=\linewidth]{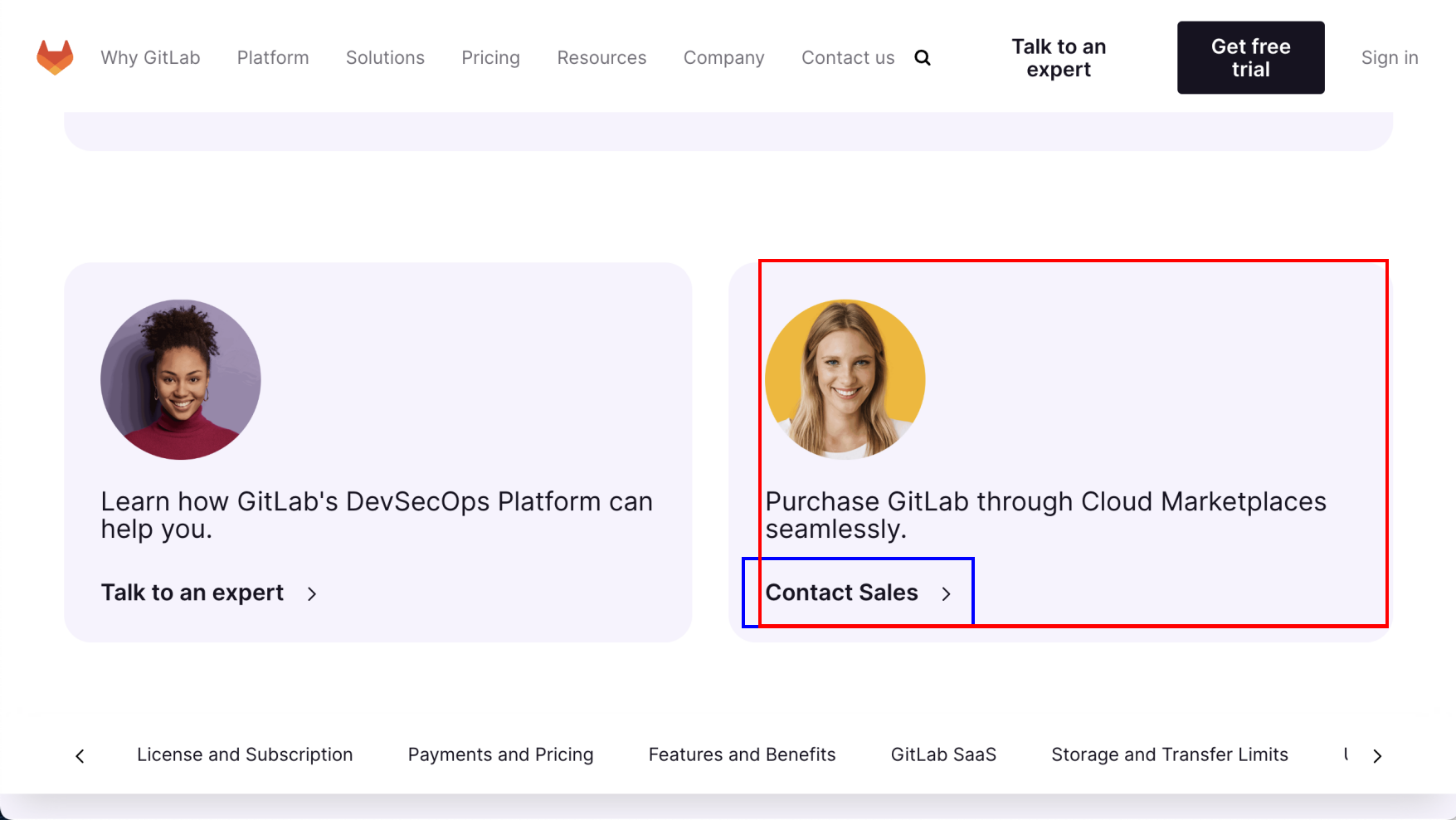}
        \caption{ScreenSpot-v2: ``contact sales'' (Greedy)}
        \label{fig:rc_case2a}
    \end{subfigure}
    \hfill
    \begin{subfigure}[t]{0.48\linewidth}
        \centering
        \includegraphics[width=\linewidth]{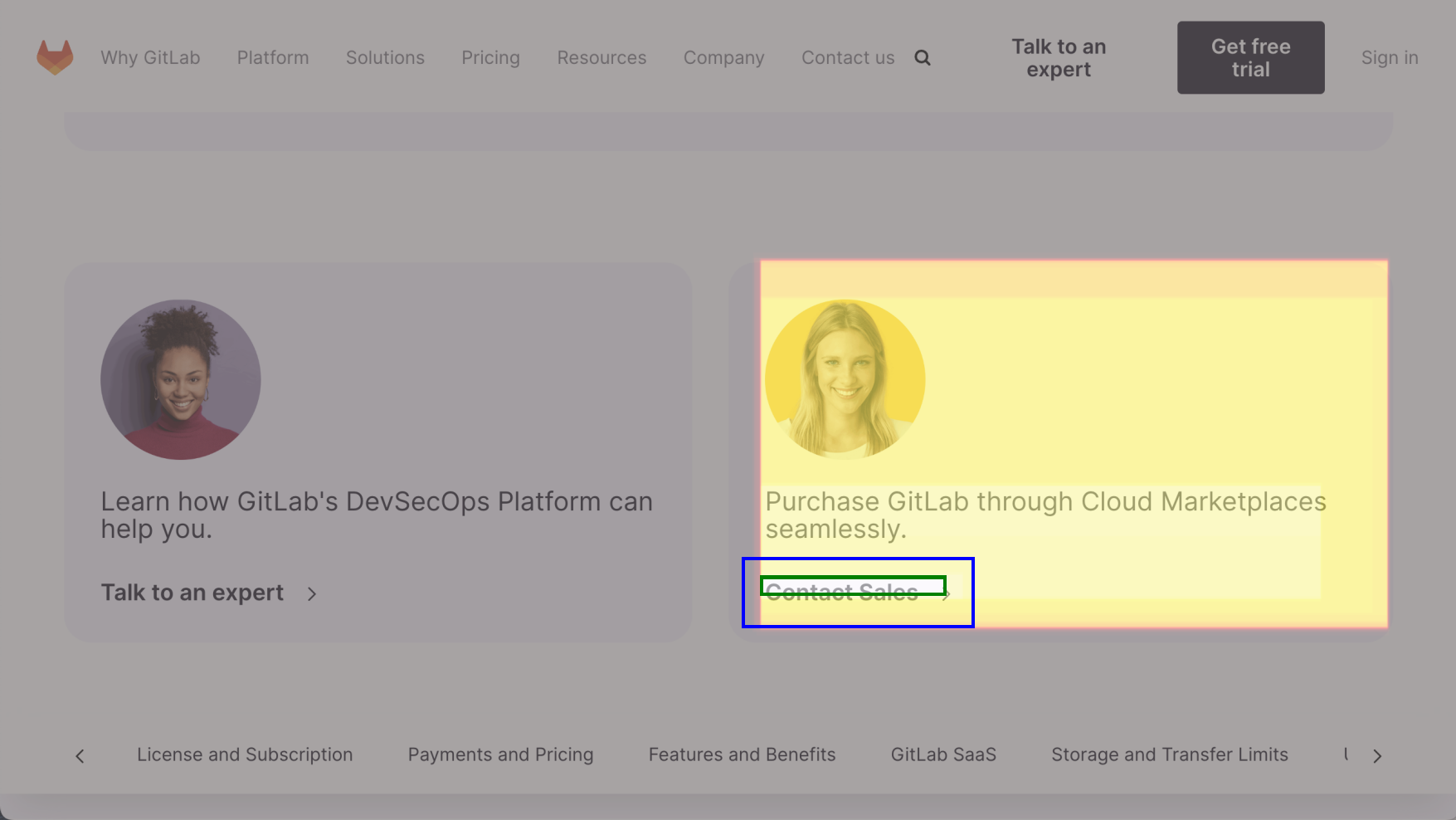}
        \caption{ScreenSpot-v2: ``contact sales'' (GUI-RC)}
        \label{fig:rc_case2b}
    \end{subfigure}

    \vspace{1em}

    \caption{
        Case studies of how GUI-RC mitigates two types of hallucinations in GUI grounding.
        In each row, the left image shows the \textbf{Greedy Decoding} result, where the blue box denotes the ground truth, and the red box denotes the model's prediction. 
        The right image shows the spatial voting heatmap obtained after applying \textbf{GUI-RC}. The brighter regions reflect higher region consistency, and the green box denotes the extracted \textbf{consensus region}.
    }
    \label{fig:rc_hallucination_cases}
\end{figure*}

To provide a more intuitive understanding on how GUI-RC mitigates two types of hallucinations, \textbf{misleading} and \textbf{biased}, during GUI grounding, we present typical failure cases drawn from the evaluation results of Qwen2.5-VL-7B-Instruct, and show how the model successfully localizes the target elements after applying GUI-RC. These examples are illustrated in Figure~\ref{fig:rc_hallucination_cases}.

\paragraph{Mitigating misleading hallucinations.} Due to the complexity of GUI layouts and the semantic similarity between UI elements, models often get confused when aligning instructions with potential target elements. As shown in Figure~\ref{fig:rc_case1a}, the instruction asks \textit{``check shoes under 50 dollars in `shop deals in fashion' part''}, but under greedy decoding, the model mistakenly selects the region of \textit{``tops under 25 dollars''}. This may be caused by a misunderstanding of the elements' semantics, leading to an incorrect match with the intended target. After applying GUI-RC, the model performs multiple samplings and constructs a spatial voting grid to identify the region with the highest prediction consistency across samples. As shown in Figure~\ref{fig:rc_case1b}, this consensus region successfully matches the ground-truth bounding box, and effectively mitigates the misleading hallucinations.

\paragraph{Mitigating biased hallucinations.} Biased hallucinations primarily stem from the granularity mismatch between pixel-level coordinate predictions and patch-level representations processed by modern visual encoders. As shown in Figure~\ref{fig:rc_case2a}, the instruction asks \textit{``contact sales''}. Although the model understands the general target location, its direct prediction encompasses the entire contact card rather than precisely identifying the \textit{``Contact Sales''} button, resulting in failed grounding. GUI-RC mitigates this issue by aggregating the regions from multiple predictions and extracting the most attention-concentrated area during sampling. As shown in Figure~\ref{fig:rc_case2b}, the consensus region precisely matches the location of the target element, enabling successful grounding and significantly improving the model’s robustness. 

\begin{table*}[t!]
\centering
\small
\begin{tabularx}{\textwidth}{lX}
\toprule
\textbf{Model} & \textbf{Prompt Template} \\
\midrule
Qwen2.5-VL & 
\texttt{You are a GUI analysis expert. Based on the screenshot, your task is to locate the UI element that best matches the instruction: '\{instruction\}'. \newline
You MUST output exactly one bounding box in the format [x1, y1, x2, y2] strictly. Do not include any explanations, descriptions, or additional text. \newline
**Output format: [x1, y1, x2, y2]**} \\
\midrule
InternVL3 & 
\texttt{SYSTEM: You are InternVL, a GUI agent. You are given a task and a screenshot of the screen. You need to perform a series of pyautogui actions to complete the task. \textbackslash n<image>\textbackslash n \newline
USER: '\{instruction\}'}. \\
\midrule
UGround & 
\texttt{Your task is to help the user identify the precise coordinates (x, y) of a specific area/element/object on the screen based on a description.\newline
- Your response should aim to point to the center or a representative point within the described area/element/object as accurately as possible.\newline
- If the description is unclear or ambiguous, infer the most relevant area or element based on its likely context or purpose.\newline
- Your answer should be a single string (x, y) corresponding to the point of the interest. \newline
Description: '\{instruction\}'}. \\
\midrule
OS-Atlas & 
\texttt{In this UI screenshot, what is the position of the element corresponding to the command '\{instruction\}' (with bbox)?} \\
\midrule
UI-TARS & 
\texttt{You are a GUI analysis expert. Based on the screenshot, your task is to identify the center point of the UI element that best matches the instruction: '\{instruction\}'. \newline
You MUST output exactly one point coordinate in the format [x, y] strictly. Do not include any explanations, descriptions, or additional text. \newline
**Output format: [x, y]**} \\
\bottomrule
\end{tabularx}
\caption{Prompt templates used for different models in GUI grounding evaluation. The \texttt{\{instruction\}} placeholder is replaced with the actual input at inference time.}
\label{tab:prompt_templates}
\end{table*}

\subsection{C. Evaluation Details}
This section provides an overview of the benchmarks, models, and prompt templates used in the evaluation of GUI grounding performance.
\subsubsection{C.1 GUI Grounding Benchmarks} 
\begin{itemize}
    \item \textbf{ScreenSpot}~\citep{cheng2024seeclickharnessingguigrounding} is a widely used benchmark for GUI grounding, containing over 600 multi-platform screenshots with 1,272 instructions across mobile, web, and desktop domains. Each sample consists of a natural language instruction, a screenshot and the corresponding target region annotated as a bounding box. It serves as a standard zero-shot evaluation dataset.
    \item \textbf{ScreenSpot-v2}~\citep{wu2024osatlasfoundationactionmodel} is an enhanced version of ScreenSpot with improved annotation quality. It contains 1,272 instructions with balanced distribution across platforms. Annotation errors from the original dataset (e.g., ambiguous or incorrect labels) were manually corrected to ensure more reliable benchmarking.
    \item \textbf{ScreenSpot-Pro}~\citep{li2024screenspot-pro} focuses on GUI grounding in high-resolution professional interfaces, covering 23 expert software applications across five professional categories and three operating systems. The dataset challenges models with complex UI layouts and fine-grained target regions. It is designed to evaluate generalization to real-world professional software.
\end{itemize}

\subsubsection{C.2 Model Details}
\begin{itemize}
    \item \textbf{Qwen2.5-VL}~\citep{bai2025qwen25vltechnicalreport} is a multimodal model from Alibaba based on the Qwen2.5 language model, extended with a visual encoder trained from scratch. It supports fine-grained localization via point or bounding box output, and is optimized for multiple multimodal tasks.
    \item \textbf{InternVL3}~\citep{zhu2025internvl3exploringadvancedtraining} is an open-source vision-language model trained jointly on vision and text modalities. Unlike adapter-based approaches, it is trained from scratch with a unified architecture and achieves strong performance on multimodal tasks.
    \item \textbf{UGround}~\citep{gou2024navigatingdigitalworldhumans} is a GUI-specific grounding model based on LLaVA. It is trained on a large-scale synthetic dataset of GUI screenshots and instructions, enabling pixel-level grounding from vision alone, without relying on DOM or accessibility metadata.
    \item \textbf{OS-Atlas}~\citep{wu2024osatlasfoundationactionmodel} is a foundation model for GUI action grounding, trained on millions of synthetic screenshots spanning five platforms. It uses unified action formats and multi-stage instruction tuning to produce robust GUI agents.
    \item \textbf{UI-TARS}~\citep{qin2025uitarspioneeringautomatedgui} is an end-to-end GUI agent that directly outputs interaction actions from screenshots and instructions. It unifies perception, grounding, and multi-step planning in a single model and is trained via interaction feedback collected from both simulated environments and real-world environments.
\end{itemize}

\subsubsection{C.3 Prompts}  
We list the prompt templates used for each model in Table~\ref{tab:prompt_templates}. For models that provide official prompt templates for GUI grounding, such as InternVL3, UGround, and OS-Atlas, we adopt the same prompts as specified in their original implementations to ensure fair comparison.

\end{document}